%% file: template.tex
\documentclass{article}

\usepackage{arxiv}

\usepackage[utf8]{inputenc} 
\usepackage[T1]{fontenc}    
\usepackage{hyperref}       
\usepackage{url}            
\usepackage{booktabs}       
\usepackage{amsfonts}       
\usepackage{nicefrac}       
\usepackage{microtype}      
\usepackage{lipsum}		
\usepackage{graphicx}
\usepackage{doi}
\usepackage[numbers]{natbib}
\usepackage{amsmath}

\input{bangla_commands}

\title{Enhancing Assamese NLP Capabilities: Introducing a Centralized Dataset Repository}

\author{ \href{https://orcid.org/0009-0007-8038-1278}{\includegraphics[scale=0.06]{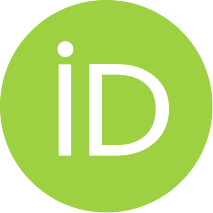}\hspace{1mm}Sagar Tamang}\thanks{
    Correspondance can be addressed to \textit{cs22bcagn033@kazirangauniversity.in}} \\
	Department of IT\\
	The Assam Kaziranga University\\
	Jorhat, India \\
	\texttt{cs22bcagn033@kazirangauniversity.in} \\
	\And
	\href{https://orcid.org/0000-0001-7809-5220}{\includegraphics[scale=0.06]{orcid.pdf}\hspace{1mm}Dr. Dibya Jyoti Bora} \\
	Department of IT\\
	The Assam Kaziranga University\\
	Jorhat, India \\
	\texttt{dibyajyotibora@kazirangauniversity.in} \\
}

\renewcommand{\shorttitle}{Enhancing Assamese NLP Capabilities: Introducing a Centralized Dataset Repository}

\hypersetup{
pdftitle={Introducing a Centralized Dataset Repository},
pdfsubject={cs.AI, cs.CL},
pdfauthor={S. Tamang, D. J. Bora},
pdfkeywords={Assamese, corpora, NLP, LLMs, repository},
}

\begin{document}
\maketitle

\begin{abstract}
This paper introduces a centralized, open-source dataset repository designed to advance NLP and NMT for Assamese, a low-resource language. The repository, available at GitHub\footnote{\url{https://github.com/indian-nlp/assamese-dataset}}, supports various tasks like sentiment analysis, named entity recognition, and machine translation by providing both pre-training and fine-tuning corpora. We review existing datasets, highlighting the need for standardized resources in Assamese NLP, and discuss potential applications in AI-driven research, such as LLMs, OCR, and chatbots. While promising, challenges like data scarcity and linguistic diversity remain. The repository aims to foster collaboration and innovation, promoting Assamese language research in the digital age.
\end{abstract}

\keywords{Assamese \and corpora \and NLP \and LLMs \and repository
}

\section{Introduction}
\label{sec:introduction}
The introduction of Large Language Models (LLMs) has led to remarkable progress in the Neural Machine Translation (NMT). However, such models require an ever-larger text corpus to train them \cite{dodge2021documentinglargewebtextcorpora}. NLP researches are basically dependent on resources such as the datatset. Low-resource languages, such as Assamese, face this challenge ever so often \cite{nlp-assamese-upos-tagged}. Thus, a centralized open-source repository of the Assamese dataset would help a community of NLP researchers and enthusiasts alike to participate in maintaining or upgrading the dataset.

Assamese language is the official language of Assam, a north-eastern state of India, with home to approximately 15 million native speakers. It is one of the scheduled languages of India. A variety of experiments and researchers are being conducted in the Assamese language. With the growing interest in research in this field, comes the requirement for datasets. Various datasets has been used to facilitate such research works, few standard but mostly nonstandard datasets are used for the tasks \cite{nlp-assamese-upos-tagged}. 

Through our research, we propose a central repository of  datasets for the Assamese language curated for the NMT Research purposes.

\section{Literature Review}
\label{sec:literature}
C. J. Kumar and S. K. Kalita (2016) developed a comprehensive dataset aimed at enhancing research in Optical Character Recognition (OCR) for the Assamese script. They created a novel offline dataset that includes a wide range of both modern and historical Assamese documents, encompassing machine-printed, handwritten, and degraded texts. The dataset was meticulously collected from various sources, ensuring a broad representation of Assamese language variations. Their paper also addressed the challenges inherent in recognizing Assamese characters, particularly in degraded and historical documents. By introducing the dataset, the authors aimed to support and drive further research in Assamese OCR, facilitating the sustainability of the language in the digital era \cite{kumarchandanpreparation}.

K. Talukdar and S. K. Sarma (2024), in their study, developed a Universal Parts of Speech (UPoS) tagged dataset for the Assamese language, which was crucial for advancing Natural Language Processing (NLP) and AI research in low-resource languages. They created a dataset comprising 283,506 tokens across 20,280 sentences by mapping the BIS tagset to the UPoS tagset. After rigorous linguistic validation, their dataset served as a gold standard resource for Assamese, supporting various NLP tasks and enabling the training of machine learning models for the language \cite{nlp-assamese-upos-tagged}.

D. Pathak et al. (2022), in their study, developed a named entity recognition (NER) dataset for the Assamese language. Their dataset consisted of around 99,000 tokens, including text from political speeches and Assamese plays. They benchmarked this dataset using various state-of-the-art NER models, achieving an 80.69\% F1-score with the MuRIL model. The annotated dataset and top-performing models were made publicly available to support further research in Assamese NLP \cite{pathak2022asnerannotateddataset}.

P. Chowdhury et al. (2023) developed an acoustic/prosodic feature-based audio dataset specifically for Assamese speech summarization. They focused on extracting prosodic features such as pitch, intensity, and sentence duration to aid in summarizing spoken Assamese content. The dataset was designed to support the development and evaluation of speech summarization models. The authors also planned to expand the dataset by adding manual transcripts and incorporating additional prosodic elements like intonation to enhance the accuracy of the summarization models \cite{chowdhury-assamese-speech-summarization}.

U. Baruah and S. M. Hazarika (2014) introduced a novel dataset of online handwritten Assamese characters, collected from various contributors using digitizers. Their dataset comprised of 8,235 samples covering numerals, basic alphabetic characters, and conjunct consonants. They detail the data acquisition process, which involved recording pen-tip movements and switching states, and evaluate preliminary classification results using support vector machines (SVMs) with different kernels. They explored three feature sets to assess classification accuracy, achieving a 99.11\% accuracy for numerals and 81.15\% for alphabetic characters with specific SVM kernels. Their dataset is the first publicly available resource for online handwritten Assamese characters, intended to support further research in handwriting recognition for this script \cite{dataset-handwritten-assamese}.

P. Choudhury et al. (2024) explored Automatic Image Captioning (AIC) for the low-resource Assamese language in their study. They examined two distinct approaches: one that used an English-pretrained model to generate English captions, which were then translated into Assamese, and another that trained a model directly on an Assamese image-caption dataset. Their research evaluated the performance of these approaches using both LSTM and transformer architectures. Their findings indicated that models trained specifically on Assamese data outperformed those relying on English-to-Assamese translation. The study demonstrated the superior efficacy of language-specific training for image captioning in low-resource languages \cite{choudhury-2024-image-captioning}.

Das and Singh (2024) examined sentiment analysis for the Assamese language, a resource-constrained language despite its official status in India. Their study utilized various lexical features from Assamese news texts and applied both machine learning and deep learning techniques. They found that combining AAV lexical features with an XGBoost classifier achieved the highest accuracy of 86.76\% using the TF-IDF approach. Their findings highlight that integrating lexical features with machine learning models significantly enhances sentiment prediction, particularly in scenarios with small datasets \cite{das-2024-emperical-study-assamese-sentiment}.

S. Sarma and N. Pathak (2023) proposed a design for an AI chatbot capable of understanding Assamese using a feedforward neural network (FFNN). They developed an Assamese language dataset and trained the model utilizing the bag-of-words method for feature extraction. The model was evaluated based on accuracy, precision, recall, and F1 score, achieving an accuracy of 89.11\%, which surpassed traditional SVM and Naive Bayes models. Their research introduced a deep-learning FFNN model for an Assamese chatbot and provided future research directions. This study is notable for its contribution to the AI Assamese chatbot domain, offering potential advancements for digital language use and inspiring further research \cite{Sarma-2023-assamese-chatbot-nn}.

The rest of the paper is organized in this manner: Section \ref{sec:methodology} outlines the methodology for obtaining the data; Section \ref{sec:usages-applications} showcases the possible use cases of the data from the repository; Section \ref{sec:challenges} delves deep into the challenges and limitations of the repository; Section \ref{sec:discussion} discusses the future scope of the research; And finally Section \ref{sec:conclusion} gives the conclusion of the research.

\section{Dataset Collection}
\label{sec:methodology}
The datasets are divided into two types, pre-training corpora and post-training corpora. As the name suggests, pre-training corpora are to be used to train the NMT Models for the pre-training steps; Similarly, the post-training corpora are to be used, though not limited to, for the fine-tuning of the NMT Models. 

As mentioned in table \ref{table:corpora}, tentative 4 pre-training corpora and 5 post-training corpora exist in the repository.

\begin{table}[h!]
\centering
\caption{Pre-training and Fine-tuning Corpora}
\begin{tabular}{|c|l|}
\hline
\textbf{Sr No} & \textbf{Name} \\ \hline
\multicolumn{2}{|c|}{\textbf{Pre-training Corpora}} \\ \hline
1 & Assamese Wikipedia v1 \\ \hline
2 & Assamese Wikipedia v2 \\ \hline
3 & CC-100 Monolingual \\ \hline
4 & The C4 Multilingual Dataset \\ \hline
\multicolumn{2}{|c|}{\textbf{Fine-tuning Corpora}} \\ \hline
5 & Assamese Sentiments Dataset \\ \hline
6 & Assamese Wikipedia Sentences Dataset \\ \hline
7 & Assamese ChatGPT Generated Dataset for Fine-Tuning \\ \hline
8 & Assamese CC-100 Multilingual Dataset for Fine-Tuning \\ \hline
9 & Assamese Poem - \bng{kobita} \\ \hline
\end{tabular}
\label{table:corpora}
\end{table}

The Wikipedia dataset is obtained through the Wikimedia dumps \cite{wikipedia_database_download} of 2021. It is separated into two versions, one for 10k lines and another for 100k lines.

CC100 corpus comprises of monolingual data for 100+ languages and also includes data in romanized language. We have given the URL only for the Assamese language \cite{lin-etal-2022-shot}.

The C4 multilingual dataset consists of more than 101 languages including the Assamese. It is a colossal, cleaned version of Common Crawl's web crawl corpus \cite{Kreutzer_2022}.

The ChatGPT generated dataset, as the name suggests, is generated by ChatGPT 3.5. It consists of 10,000 lines of JSONL file, with "word" and "sentence" keys, making it specially desirable for Fine Tuning purposes.

The finetuning version of CC-100 Multilingual is just a jsonl converted, making it preferable for fine tuning.

The last dataset at disposal as of today is the Assamese poem \text{\bng{kobiTaa}}3 which is scraped from \url{https://www.arambhani.com/}{https://www.arambhani.com/}.

\begin{figure}
    \centering
    \includegraphics[width=0.8\linewidth]{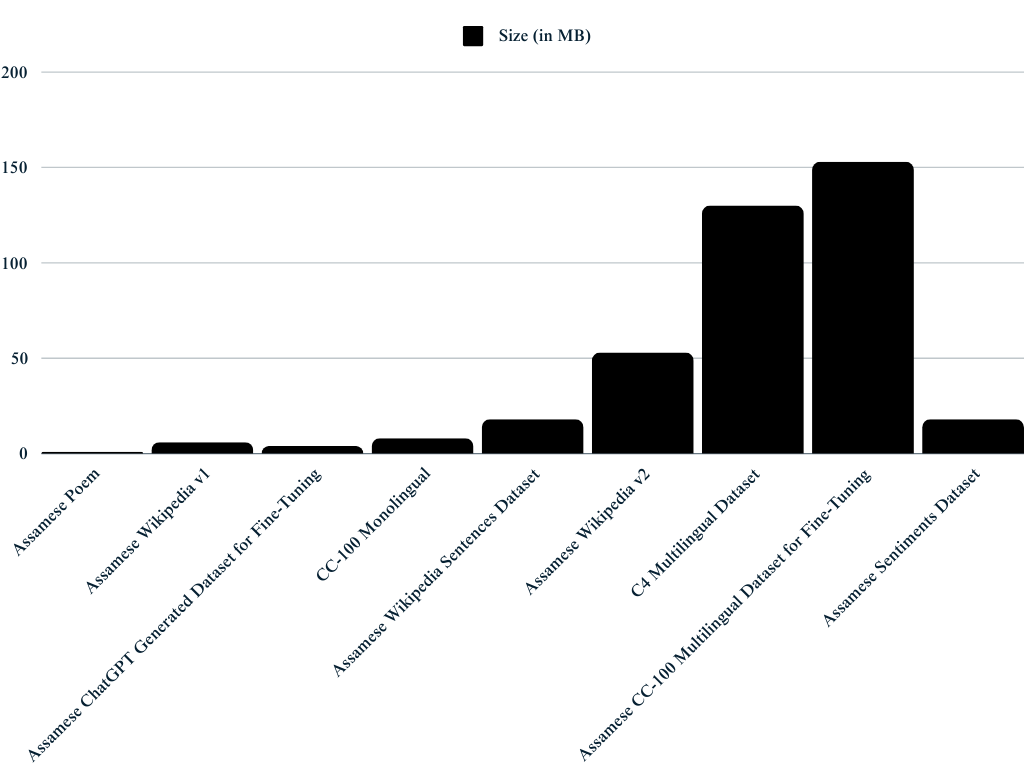}
    \caption{Size of Each Corpora in the Dataset}
    \label{fig:figure-1}
\end{figure}

Figure \ref{fig:figure-1} illustrates the size of each corpora in the dataset. As we can observe, the Assamese CC-100 Multilingual, C4 Multilingual Dataset, and Assamese Wikipedia v2 comprise the majority of the storage usage.

\section{Usage and Applications}
\label{sec:usages-applications}

The centralized dataset repository for the Assamese language presents a broad spectrum of usage and applications, primarily in the domains of Natural Language Processing (NLP), Machine Translation, and AI-driven linguistic research. One of the primary applications is in the development and refinement of Large Language Models (LLMs) specific to Assamese, which can significantly enhance the accuracy and reliability of tasks such as machine translation, sentiment analysis, and named entity recognition (NER). Researchers can leverage these datasets for training deep learning models, thereby enabling more sophisticated AI systems capable of understanding and generating text in Assamese.

Moreover, the repository can be utilized for developing domain-specific applications such as Optical Character Recognition (OCR) systems, which are essential for digitizing historical and contemporary Assamese documents. This can facilitate the preservation of Assamese literary and cultural heritage in the digital age. The dataset is also instrumental in creating chatbots and virtual assistants that communicate in Assamese, thereby expanding the reach of digital services to Assamese speakers.

Educational institutions and linguists can use the repository for language teaching tools and automatic grading systems that can evaluate Assamese text with higher precision. Additionally, the dataset can support sentiment analysis applications for monitoring and analyzing public opinion in Assamese news and social media, aiding in better understanding societal trends.

Finally, the repository offers significant potential for use in cross-lingual applications, where models trained on Assamese data can be adapted or translated into other languages, thus fostering multilingual NLP research and bridging the gap between high-resource and low-resource languages.

\section{Challenges and Limitations}
\label{sec:challenges}

Despite the promising prospects, there are several challenges and limitations associated with the creation and utilization of a centralized Assamese dataset repository. One of the primary challenges is the inherent scarcity of high-quality, annotated Assamese text data, which limits the ability to train robust NLP models. The lack of standardized datasets poses difficulties in ensuring consistency across different research projects, making it challenging to compare results or replicate studies.

Another significant challenge is the diversity within the Assamese language itself, which includes various dialects and scripts. This linguistic variability can complicate dataset collection, annotation, and model training, as models may struggle to generalize across different forms of Assamese.

The availability of computational resources also poses a limitation, particularly for researchers and institutions with limited access to high-performance computing infrastructure. Training LLMs and other complex models on large datasets requires substantial computational power, which may not be readily available in low-resource settings.

Furthermore, there are ethical considerations related to data privacy and consent, especially when dealing with sensitive or personally identifiable information in the datasets. Ensuring the ethical use of data and protecting the privacy of individuals whose data may be included in the corpus is a critical concern that must be addressed.

Finally, there is the challenge of maintaining and updating the dataset repository. As the Assamese language continues to evolve, and as new data sources become available, the repository will need to be regularly updated to remain relevant and useful. This requires ongoing collaboration and support from the research community and other stakeholders.

\section{Discussion}
\label{sec:discussion}

The development of a centralized dataset repository for the Assamese language is a crucial step towards advancing NLP research and AI applications for low-resource languages. This initiative addresses a critical gap in the availability of standardized, high-quality datasets, which are essential for training effective models in the Assamese language. By providing a curated and accessible repository, the research community can foster greater collaboration and innovation in Assamese NLP.

However, the challenges and limitations outlined in Section \ref{sec:challenges} highlight the need for a strategic approach to dataset collection, annotation, and maintenance. It is essential to develop methodologies that can efficiently capture the linguistic diversity of Assamese while ensuring data quality and consistency. The role of community-driven efforts cannot be overstated, as a collective approach can help mitigate some of the resource constraints and ensure the sustainability of the repository.

The discussion also extends to the ethical implications of dataset usage. Researchers must be mindful of the privacy concerns and ethical considerations associated with the data, especially when deploying models in real-world applications. Transparency in data usage and adherence to ethical standards will be critical in gaining public trust and ensuring the responsible use of AI technologies in the Assamese language context.

Looking forward, the repository could serve as a model for similar initiatives in other low-resource languages, promoting linguistic diversity and inclusivity in the global NLP landscape. By building robust and accessible datasets, we can help ensure that all languages, regardless of their resource status, are represented in the digital age.

\section{Conclusion}
\label{sec:conclusion}

In conclusion, the introduction of a centralized dataset repository for the Assamese language marks a significant milestone in the field of NLP and AI research. This repository not only addresses the current gaps in data availability for Assamese but also provides a foundational resource that can drive innovation and enhance the performance of various language technologies. By offering a curated collection of pre-training and fine-tuning corpora, the repository enables researchers to develop more accurate and culturally relevant models for Assamese, thereby contributing to the preservation and promotion of the language in the digital era.

While challenges related to data scarcity, linguistic diversity, and computational resources remain, the potential benefits of this repository far outweigh the limitations. The collaborative efforts of the research community, combined with ongoing advancements in AI, will be key to overcoming these challenges and ensuring the continued growth and relevance of Assamese NLP.

Ultimately, this repository serves as a critical enabler for future research and applications, fostering greater inclusion of Assamese in the global AI ecosystem. As the field of NLP continues to evolve, this repository will play a pivotal role in ensuring that the Assamese language remains vibrant and accessible in the digital world.

\section*{Acknowledgments}
\label{sec:acknowledgment}
We would like to thank \textit{The Assam Kaziranga University} for providing us with the necessary resources to conduct this research.

\end{document}

%% file: bangla_commands.tex
\def\bng{\bngx}

\font\bngx=bang10

\def\*#1*#2{o\null{#2}{#1}}

\def\sh#1{\setbox0=\hbox{#1}%
     \kern-.02em\copy0\kern-\wd0
     \kern.04em\copy0\kern-\wd0
     \kern-.02em\raise.0433em\box0 }